\documentclass{article}

\usepackage[final]{neurips_2023_ml4ps}

\usepackage[margins]{trackchanges}
\addeditor{YM}

\usepackage[utf8]{inputenc} 
\usepackage[T1]{fontenc}    
\usepackage{hyperref}       
\usepackage{url}            
\usepackage{booktabs}       
\usepackage{amsfonts}       
\usepackage{nicefrac}       
\usepackage{microtype}      
\usepackage{xcolor}         
\usepackage{graphicx}
\usepackage{hyperref}
\usepackage{subfig}

\title{Leveraging Deep Learning for Physical Model Bias of Global Air Quality Estimates}

\author{%
  Kelsey Doerksen\thanks{Use footnote for providing further information
    about author (webpage, alternative address)---\emph{not} for acknowledging
    funding agencies.} \\
  Department of Computer Science\\
  OATML\\
  NASA Jet Propulsion Lab \\
  University of Oxford\\
  \texttt{kelsey.doerksen@cs.ox.ac.uk} \\
  \And
  Yuliya Marchetti \\
  Jet Propulsion Lab \\
  California Institute of Technology \\
  \And
  Kevin Bowman \\
  Jet Propulsion Laboratory\\
  California Institute of Technology\\
  \And
  Steven Lu \\
  Jet Propulsion Laboratory\\
  California Institute of Technology\\
  \And
  James Montgomery \\
  Jet Propulsion Laboratory\\
  California Institute of Technology\\
  \And
  Yarin Gal \\
  OATML \\
  University of Oxford\\
  \And
  Freddie Kalaitzis \\
  OATML \\
  University of Oxford\\
  \And
  Kazuyuki Miyazaki \\
  Jet Propulsion Laboratory\\
  California Institute of Technology
}

\begin{document}

\maketitle

\begin{abstract}
Air pollution is the world’s largest environmental risk factor for human disease and premature death, resulting in more than 6 million premature deaths in 2019. Currently, there is still a challenge to model one of the most important air pollutants, surface ozone (O3), particularly at scales relevant for human health impacts, with the drivers of global ozone trends at these scales largely unknown, limiting the practical use of physics-based models. We employ a 2-D Convolutional Neural Network (CNN)-based U-Net architecture that estimates surface ozone MOMO-Chem model residuals, referred to as model bias. We demonstrate the potential of this technique in North America and Europe, highlighting its ability better to capture physical model residuals compared to a traditional machine learning method.  We assess the impact of incorporating land use information from high-resolution satellite imagery to improve model estimates. Importantly, we discuss how our results can improve our scientific understanding of the factors impacting ozone bias at urban scales that can be used to improve environmental policy.
\end{abstract}

\section{Introduction}
The 2017-2027 Decadal Survey for Earth Science and Applications from Space stated it's priority to define "What processes determine the spatio-temporal structure of important air pollutants and their concomitant adverse impact on human health, agriculture, and ecosystems?" [1]. However, modeling air pollution to date has been difficult due to the complex interconnected relationships between atmospheric chemistry, emissions, planetary boundary layer dynamics and unknown non-linear processes. Recently, the Multi-mOdel Multi-cOnstituent Chemical data assimilation (MOMO-Chem) for tropospheric chemical reanalysis was introduced to account for physical model error in transport and chemistry through data assimilation analyses [2]. Though MOMO-Chem has made significant progress in reproducing large-scale ozone estimates, there is a gap in finer-scale ozone analysis and drivers of physical model ozone bias relevant for human health assessments.

In this work, we investigate the integration of high-resolution satellite data products with the MOMO-Chem physical model to train a 2-D U-Net to predict daily 8-hour surface ozone bias across North America and Europe. \textbf{Physical model bias} is defined as the difference, or systematic error, between a physical model’s output and ground truth observation for a given target variable i.e. surface ozone. We show that the addition of land use information improves our predictions of model bias and that our deep learning model can better capture bias extremes over traditional machine learning (ML) methods. This work provides a first application of Deep Learning for predicting and diagnosing MOMO-Chem physical model residuals.

\section{Background and Related Work}
At the Earth's surface, ozone is an air pollutant formed through chemical reactions in the atmosphere when ultraviolet radiation from the sun interacts with nitrogen oxides and volatile organic compounds [3]. The MOMO-Chem model is a state-of-the-art data assimilation framework used to estimate surface ozone, but suffers from large systematic estimation errors (bias) due to insufficient information from the current observing systems, leading to a limited understanding of air quality and its health impacts. In MOMO-Chem, bias can be driven by a mixture of poorly resolved and unresolved processes including atmospheric chemistry, planetary boundary layer dynamics and emissions from human behaviour. 

\subsection{Machine Learning for Air Quality Physical Model Bias}
Deep Learning can be leveraged to identify mechanisms driving near-surface pollution and correct for their impact on air quality predictions, thereby improving physical models. Recent work has been developed to capture physical model bias for climate, weather and Earth system models with deep learning in [4, 5, 6], and traditional Random Forest (RF) ML techniques have been applied to model ozone concentration bias of the GEOS-Chem Chemical Transport Model in China in 2018 [7]. Our work proposes that a U-Net-based architecture is better suited to capture ozone bias than a Random Forest, based on its ability to capture spatial relationships between neighbouring pixels, and through a combination analysis of RF and U-Net results, a clearer picture of the drivers of surface ozone bias can begin to be uncovered.

\section{Dataset}
\begin{figure}[t]
\begin{center}
\includegraphics[scale=0.14]
{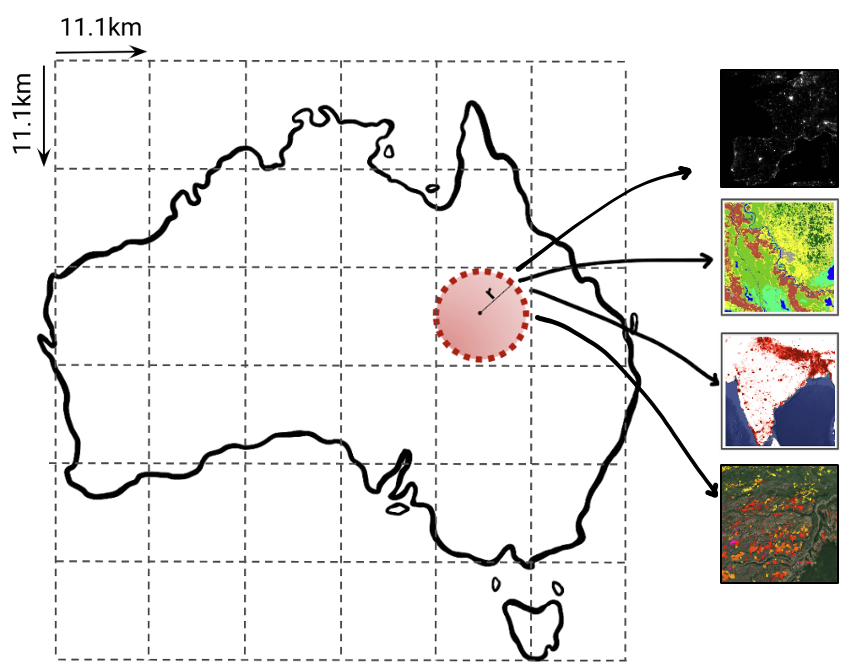}
\caption{airPy GEE extraction over Australia. Top to bottom datasets: VIIRS nightlight, MODIS landcover, World Population and MODIS Burnt Pixel data}
\label{fig:airpy}
\end{center}
\end{figure}

\textbf{Input} \; \; MOMO-Chem generates 126, 2-hourly estimates of ozone parameters and atmospheric chemical composites, covering a 160x120 point latitude, longitude grid of the Earth at a 1x1 degree per pixel resolution. These features were down-selected to an emulator version of 16 of the top-ranked features by importance from RF experiments and domain expert insight retaining: ammonia, dimethyl sulfide, nitric acid, carbon monoxide, bromine nitrate, temperature, nitrogen dioxide, peroxyacetyl nitrate, chemical productions of hydrogen oxide radicals, surface pressure, hydrogen superoxide, 1-Pentyne, sulfur dioxide, hydroxide clear-sky longwave radiation flux at surface and clear string outgoing longwave radiation to space. Land use data was extracted and processed into an ML-ready format using the airPy package (see \ref{airpy}) and includes the mode, variance, and percent coverage per land class per grid from the MODIS Land Cover Yearly product and the variance, maximum, minimum and average from the GPWv411 Population Density product, encompassing 23 features [8, 9].

\begin{figure}
  \centering
  \subfloat[a][Europe]{\includegraphics[scale=0.26]{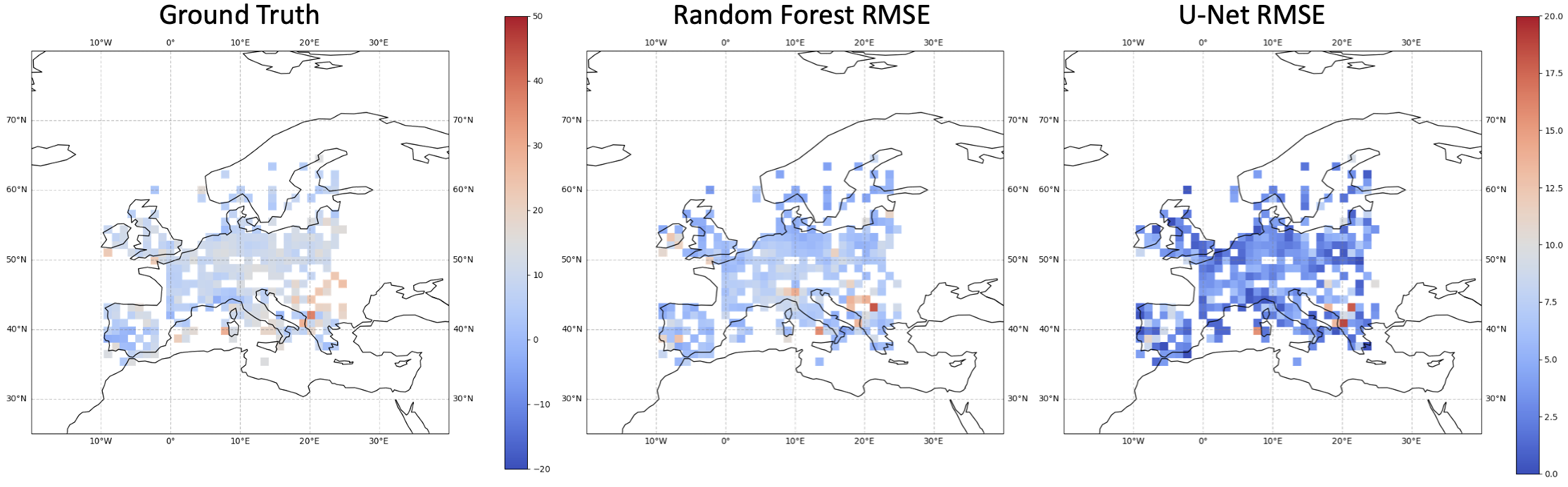} \label{fig:EU_RMSE}} \\
  \subfloat[b][North America]{\includegraphics[scale=0.26]{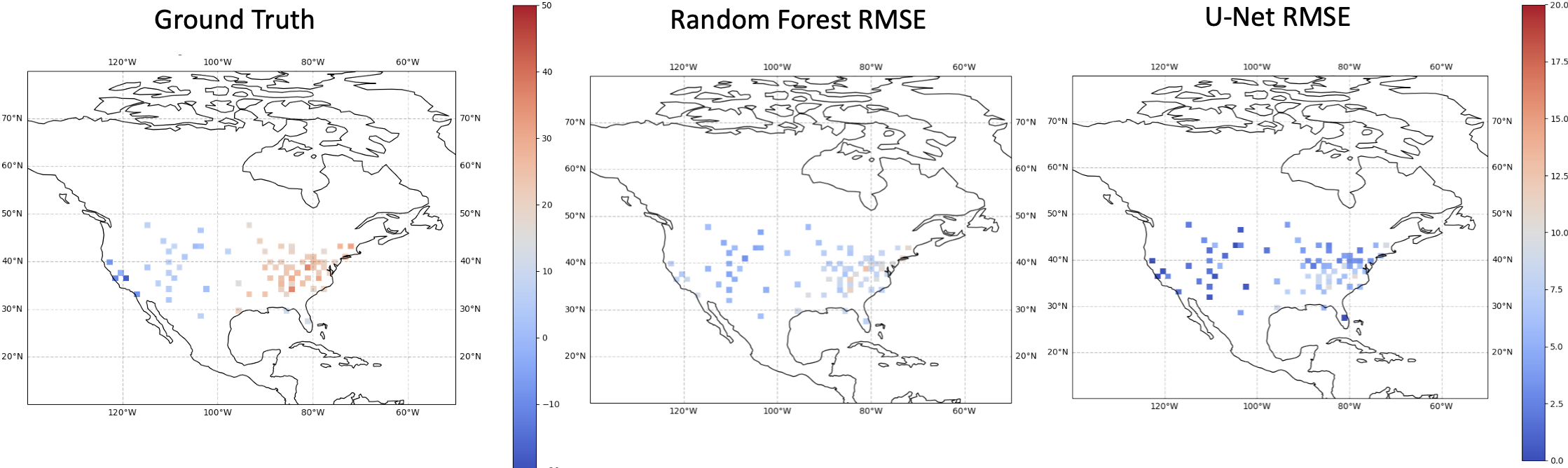} \label{fig:NA_RMSE}}
  \caption{RF baseline vs Deep Learning model average RMSE over Europe and North America. Locations are shown where ground truth data is available from the TOAR network.} \label{fig:NA_RMSE}
\end{figure}

\textbf{Ground Truth} \; \; Ground truth surface ozone data is provided by the Tropospheric Ozone Assessment Report (TOAR) database which contains one of the world’s largest collections of near-surface ozone measurements [10]. Though TOAR is the most sophisticated global ground truth network for surface ozone, the majority of stations are located across North America and Europe, with some coverage over Asia, and virtually no coverage across the Global South. This limitation restricts our study to focus on North America and Europe, with the hope that the ongoing TOAR Phase-II will provide additional global coverage and motivation for future work to extend to regions with very sparse ground truth. Bias in the context of this work is the difference between the MOMO-Chem daytime 8-hour average surface ozone output and the ground-based observation of daytime 8-hour average surface ozone from TOAR.

\section{Methodology}
\subsection{Experimental Setup and Model Architecture}
Individual models are trained for North America and Europe for the Summer season (June-August) respectively. We use a CNN-based model inspired by the U-Net architecture that accounts for spatial context in the data during training. The deep learning model is compared to a RF baseline. The North American extent experiments train with multi-channel images (arrays) of size 31x49 covering latitude, longitude ranges of (20, 55) and (-125, -70) respectively, and the European extent experiments train with multi-channel images of size 27x31 that cover latitude, longitude ranges of (35, 65), (-10, 25) respectively, matched to the MOMO-Chem grid resolution. Models are trained on two feature-space configurations (number of channels) to compare the performance with and without land use information; Experiment 1 uses 16 MOMO-Chem features and Experiment 2 uses 16 MOMO-Chem features plus 23 GEE MODIS and population data. Inputs are first normalized using z-score normalization to improve training.

We adapted the U-Net architecture to include dropout after each of the 2D convolutions in the Double Convolution module which was found to improve performance over batch normalization regularization for this application [11]. We train with Adam optimization and a weight decay of 1e-3, constant learning rate of 1e-2, training for 200 epochs and dropout rate of 0.1.

\subsection{Integrating Land Use Information from Satellite Data with airPy}\label{airpy}
airPy was developed to extract high-resolution surface information from Google Earth Engine (GEE) and compute relevant metrics for air quality studies for any location on the Earth for use in ML models and other statistical analysis. For a given latitude, longitude point and specified area of interest buffer extent, airPy extracts land surface data for the specified GEE product and calculates relevant statistical features that can match any grid resolution (Figure \ref{fig:airpy}). To support open science, airPy is open-sourced and is available at: \href{https://anonymous.4open.science/r/airPy-60CD}{airPy}.

\section{Results}
\subsection{Random Forest vs U-Net}
Figure \ref{fig:NA_RMSE} showcases the superior performance of the U-Net model, on average, of capturing MOMO-Chem bias across Europe and North America over the RF baseline between June-August 2016. Intuitively this makes sense, as the CNN-based model includes spatial context during training, and supports our hypothesis to that including this context is valuable to better capture surface ozone.

\begin{figure}[t]
\begin{center}
\includegraphics[scale=0.35]{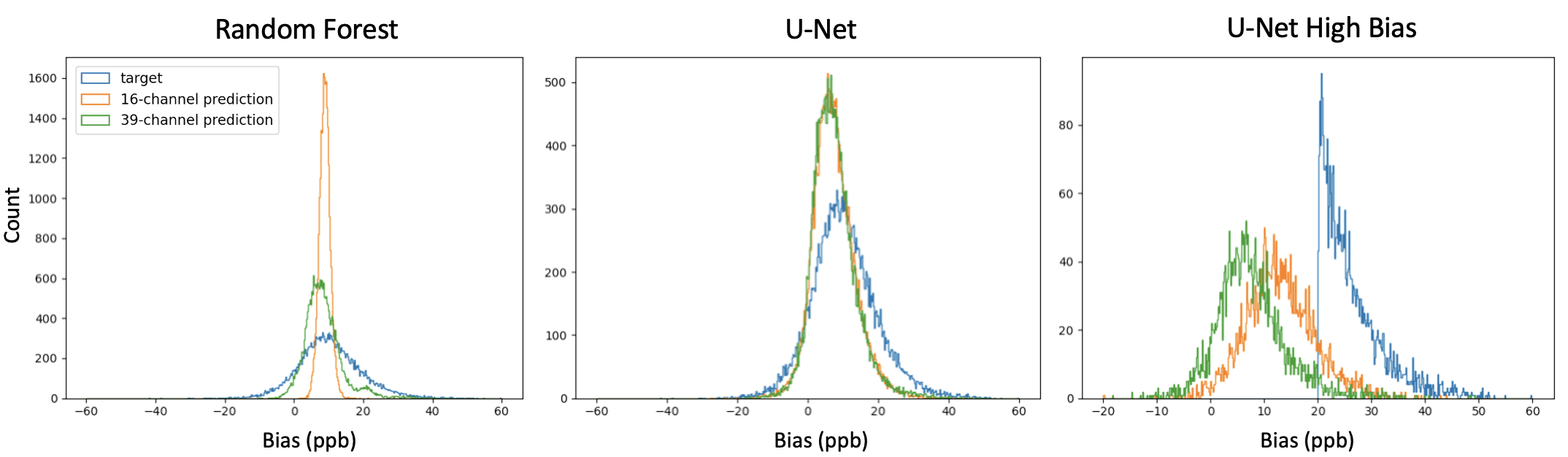}
\caption{From left to right: Bias prediction histograms over Europe for Random Forest, U-Net, and U-Net for bias values greater than 20 parts per billion (ppb).}
\label{fig:rf_unet_bias_hist}
\end{center}
\end{figure}

\subsection{Incorporating Land Use Information}
Observing the impact of adding land use data into our feature space, Figure \ref{fig:rf_unet_bias_hist} supports our hypothesis that including information derived from high-resolution satellite imagery improves model performance for the RF in capturing bias, and in particular bias extremes, over Europe. Interestingly however, this is not the case with the U-Net model, where the 39-channel feature space predicts closer to the mean of the bias distribution, in particular for cases of high bias greater than 20 ppb. These results are consistent with the North America experiments. The MODIS data is a yearly product and population data from a 5-year census, and it is possible that this temporal cadence is too coarse to improve bias estimation with the U-Net. Further investigation into this phenomena is ongoing and future work will focus on imbalanced regression and extreme value prediction problems particular to sparse data to support our investigation into driving factors of model bias.

\section{Conclusion}
In this work, we have shown for the first time the capability of deep learning to estimate physical model bias of surface ozone for the MOMO-Chem framework. The U-Net-based model outperforms the RF baseline for both Europe and North America experiments. Land use information extracted from high-resolution satellite data improves the RF model in capturing bias, but does not show improvement for the U-Net in capturing bias extremes, with further investigations ongoing in this direction. To provide additional value to the scientific community, future work will integrate uncertainty quantification methodology into our model to provide pixel-wise uncertainty estimates for predictions and explore deep learning explainability metrics including SHAP [12].

\section{Broader Impact}
This work serves as a first step to leveraging deep learning to estimate surface ozone bias with the objective to improve the MOMO-Chem framework for capturing chemical transport processes as well as integrating additional surface and human activity information into the understanding of the bias. A better prediction of the drivers of surface ozone and ozone bias are integral to correct for their impact on air quality estimates to make informed decisions to reduce global air pollution and its adverse health impacts. The development of tools like airPy will reduce computational barriers for the science community in leveraging Earth Observation data that can extend beyond air quality studies to other Earth Science applications.

\section{Acknowledgements}
KD acknowledges funding from EPSRC Centre for Doctoral Training in Autonomous Intelligent
Machines and Systems (Grant No. EP/S024050/1).

\section*{References}
{
\small

[1] E. National Academies of Sciences and Medicine, “Thriving on our changing planet: A decadal strategy for earth observation from
space.,” Washington, DC: The National Academies Press., 2018.

[2] K. Miyazaki, K. W. Bowman, K. Yumimoto, T. Walker, and K. Sudo,
“Evaluation of a multi-model, multi-constituent assimilation frame-
work for tropospheric chemical reanalysis,” Atmos. Chem. Phys,
vol. 20, no. 931-967, 2020.

[3] Duncan, Bryan. "Surface-Level Ozone." Air Quality Observations from Space, 21 Apr. 2023, airquality.gsfc.nasa.gov/.

[4] Hess, Philipp, Stefan Lange, and Niklas Boers. "Deep Learning for bias-correcting comprehensive high-resolution Earth system models." arXiv preprint arXiv:2301.01253 (2022).

[5] Wang, F., Tian, D., and Carroll, M.: Customized deep learning for precipitation bias correction and downscaling, Geosci. Model Dev., 16, 535–556, https://doi.org/10.5194/gmd-16-535-2023, 2023.

[6] Laloyaux, P., Kurth, T., Dueben, P. D., and Hall, D. (2022). Deep learning to estimate model biases in an operational NWP assimilation system. Journal of Advances in Modeling Earth Systems, 14, e2022MS003016. https://doi.org/10.1029/2022MS003016

[7] Diagnosing the Model Bias in Simulating Daily Surface Ozone Variability Using a Machine Learning Method: The Effects of Dry Deposition and Cloud Optical Depth Xingpei Ye, Xiaolin Wang, and Lin Zhang. Environmental Science \& Technology 2022 56 (23), 16665-16675 DOI: 10.1021/acs.est.2c05712

[8] D. Sulla-Menashe and M. A. Friedl, “Mcd12q1 modis/terra+aqua land cover type yearly l3 global 500m sin grid v006,” NASA EOSDIS Land Processes DAAC.

[9] C. for International Earth Science Information Network CIESIN Columbia University, “Gridded population of the world, version 4 (gpwv4): Population density,” New York: NASA Socioeconomic Data and Applications Center (SEDAC), 2018

[10] Schultz, M. G., et al.: Tropospheric Ozone Assessment Report: Database and Metrics Data of Global Surface Ozone Observations, Elem. Sci. Anth., 5, 58, https://doi.org/10.1525/elementa.244, 2017.

[11] Ronneberger, Olaf, Philipp Fischer, and Thomas Brox. "U-net: Convolutional networks for biomedical image segmentation." Medical Image Computing and Computer-Assisted Intervention–MICCAI 2015: 18th International Conference, Munich, Germany, October 5-9, 2015, Proceedings, Part III 18. Springer International Publishing, 2015.

[12] Lundberg, S. and Lee, S.-I., “A Unified Approach to Interpreting Model Predictions”, <i>arXiv e-prints</i>, 2017. doi:10.48550/arXiv.1705.07874.


\end{document}